\documentclass[10pt, a4paper]{article}

\usepackage{lrec-coling2024} 
\usepackage{amsmath,amssymb,amsfonts}
\usepackage{multirow}
\usepackage{booktabs}

\title{Arbitrary Time Information Modeling via Polynomial Approximation for Temporal Knowledge Graph Embedding}
 
\name{
Zhiyu Fang,
Jingyan Qin$^{\ast}$ \thanks{*Corresponding author},
Xiaobin Zhu,
Chun Yang,
Xu-Cheng Yin
}
\address{
    School of Computer \& Communication Engineering\\
    University of Science and Technology Beijing, Beijing, China\\
    \{mr.fangzy, qinjingyanking\}@foxmail.com, \{zhuxiaobin, chunyang, xuchengyin\}@ustb.edu.cn
}

\abstract{
Distinguished from traditional knowledge graphs (KGs), temporal knowledge graphs (TKGs) must explore and reason over temporally evolving facts adequately. However, existing TKG approaches still face two main challenges, i.e., the limited capability to model arbitrary timestamps continuously and the lack of rich inference patterns under temporal constraints. In this paper, we propose an innovative TKGE method (PTBox) via polynomial decomposition-based temporal representation and box embedding-based entity representation to tackle the above-mentioned problems. Specifically, we decompose time information by polynomials and then enhance the model's capability to represent arbitrary timestamps flexibly by incorporating the learnable temporal basis tensor. In addition, we model every entity as a hyperrectangle box and define each relation as a transformation on the head and tail entity boxes. The entity boxes can capture complex geometric structures and learn robust representations, improving the model's inductive capability for rich inference patterns. Theoretically, our PTBox can encode arbitrary time information or even unseen timestamps while capturing rich inference patterns and higher-arity relations of the knowledge base. Extensive experiments on real-world datasets demonstrate the effectiveness of our method.
\\ \newline  }

\begin{document}

\maketitleabstract

\section{Introduction}
Knowledge Graphs (KGs) are widely used in question answering, information retrieval, and recommender systems by representing human-summarized knowledge via multi-relational graphs \cite{hu2022global}. KGs usually can be viewed as a collection of facts in triple form $(h, r, t)$, which represent head entity $h$ is related to tail entity $t$ by relation $r$. However, many facts in the real-world are time-sensitive, making the triples of (static) KGs cannot describe the dynamic evolution of facts over time. For example, \textit{the president of the USA is Barack Obama only for the period 2009-2017 and is Donald Trump only for the period 2017-2021.} Therefore, Temporal Knowledge Graphs (TKGs), which introduce the timestamp to expand the fact into a quadruple form $(h, r, t, \tau)$, have recently drawn growing attention from both academic and industrial communities. 
\vspace{0.5em}

To effectively represent temporal information and construct a complete knowledge graph, Temporal Knowledge Graph Embedding (TKGE) methods usually learn low-dimensional representations of entities and relations under temporal constraints and predict missing triple links. A popular strategy in TKGE directly treats time information as the feature equivalent to entities or relations, transforming various existing KGE methods into TKGE. TComplex \cite{TComplex:conf/iclr/LacroixOU20} expands the entity-relation third-order tensor of ComplEx \cite{ComplEx:conf/icml/TrouillonWRGB16} into an entity-relation-time fourth-order tensor via canonical polyadic decomposition. Under the setting of the original KGE model, these KGE-based extension methods simply model time information into entities or relations and cannot effectively model temporal characteristics. To capture rich time information, another strategy of TKGE has been proposed. These methods typically exhibit good temporal generalization due to clever network architectures crafted for integrating time information. HyTE \cite{HyTE:conf/emnlp/DasguptaRT18} proposes hyperplane-based TKG embedding to view the TKG as a collection of KGs embedded within different temporal hyperplanes. ATISE \cite{ATiSE} utilizes the theory of additive time series decomposition to represent entities and relations as time series features that can be decomposed into trend, seasonal, and random components. DyERNIE \cite{DyERNIE:conf/emnlp/HanCMT20} defines the interaction between velocity vectors and times in the tangent space to describe the dynamic evolution of entities. Although these methods consider the impact of temporal characteristics on entity-relation pairs, they lack the capability to model arbitrary timestamps continuously.

Generally, existing TKGE methods use simple embedding vectors and the parallelogram rule in Euclidean space for feature representation, capturing limited structural information for the knowledge graph. This results in the restricted reasoning capability of the model, especially for inference patterns under temporal constraints. TTransE \cite{TTransE:conf/www/LeblayC18} employs the parallelogram rule in Euclidean space to measure the score of entity-relation pairs, which makes it unable to capture symmetry pattern. RotateQVS \cite{RotateQVS:conf/acl/ChenWLL22} defines the rotation of the entity around the time axis in quaternion vector space and calculates the score via the distance of the complex vectors, causing it cannot capture hierarchical pattern.

\begin{figure*}
\centering
\includegraphics[scale=1.2]{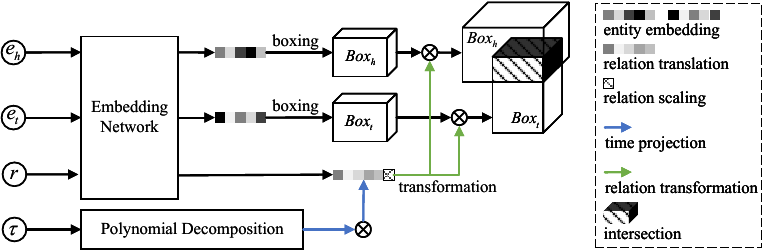}
\caption{the overall of our proposed PTBox. $e_h$, $e_t$, $r$ and $\tau$ denote head entity, tail entity, relation and timestamp for one quadruple fact $(h,r,t,\tau)$, respectively. $\otimes$ denotes Hadamard product. Our model obtains the time representation via the polynomial decomposition mechanism and represent the probability of the fact being established by the intersection between entity boxes.}
\label{fig:overall}
\end{figure*}

To address the above problems, we propose an innovative TKGE method based on polynomial approximation, namely PTBox\footnote{We release our code at\\ https://github.com/seeyourmind/PTBox}. Our method comprises two modules: polynomial decomposition-based temporal representation and box embedding-based entity representation. Specifically, we leverage the polynomial approximation theory to decompose the embedding of any given time point into a product of a coefficient vector and a learnable feature tensor. In this way, our approach allows easily for representation of arbitrary time points, even those that have never been encountered before. Due to the inherent properties of box-type geometric embedding, it can represent varied relations with respect to transitivity and is closed under intersection. Hence, we further represent the head and tail entities using box embeddings to enhance the rigid inference ability of PTBox. Empirically, we conduct detailed experimental evaluations over two popular TKGE benchmarks and prove our method can capture time-evolving information. Moreover, we analyze the learned box embeddings and show the abilities of our PTBox for modeling various relation patterns, including temporal evolution.

In summary, our main contributions are three-fold:
\begin{itemize}
\item We propose an innovative temporal knowledge graph embedding method, namely PTBox. Experimental results verify the state-of-the-art performance of our method on two publicly available datasets.
\item Proposing an interpretable time representation method that decomposes time information by polynomial approximation theory to flexibly represent arbitrary timestamp.
\item Proposing a box-embedding-based entity representation method that effectively represents calibrated probability distributions and learns rigid inference patterns.
\end{itemize}

\section{Related Work}
\subsection{Knowledge Graph Embedding}
Real-world knowledge graphs are usually incomplete, necessitating completion techniques to enable their effective application in downstream tasks. Knowledge graph embedding (KGE), as the popular static knowledge graph completion technique, employs embeddings to represent entities and relations. By learning scores for all possible facts, KGE methods predict missing links, e.g. $(the\ Beatles,\ genre,\ ?)$. As an active research area, numerous methods have been proposed to address the challenges of representing and completing knowledge graphs. These KGE methods can be broadly categorized into several key approaches: translation-based methods, tensor-based methods, and neural network-based methods.

Translation-based methods define scoring functions based on the translation between entities and relations. The classical model TransE \cite{TransE:conf/nips/BordesUGWY13} utilizes the parallelogram rule in vector space to define the scoring function, assuming that adding the head entity vector and the relation vector should be as close as possible to the tail entity vector. Subsequently, several improved models in the Trans-series have been proposed, such as TransH \cite{TransH:conf/aaai/WangZFC14}, TransN \cite{TransN:conf/sigir/WangC18}, TransC \cite{TransC:conf/emnlp/LvHLL18}. To address the contradiction between the optimization and regularization in the TransE, TorusE \cite{TorusE:conf/aaai/EbisuI18} introduces the Lie Group to learn embedding representations in a torus space. To model rich inference patterns (e.g., symmetry/antisymmetry, inversion, and composition), RotatE \cite{RotatE:conf/iclr/SunDNT19} defines each relation as a rotation in the complex vector space from the head entity to the target entity.

Tensor-based models treat the knowledge graph as a multi-dimensional tensor and aim to factorize this tensor to learn entity and relation embeddings. RESCAL \cite{RESCAL:conf/icml/NickelTK11} factorizes the tensor into low-rank matrices, capturing the interactions between entities and relations. DistMult \cite{DistMult:journals/corr/YangYHGD14a} restricts the representation matrix of relations to diagonal matrices, greatly reducing model complexity. ComplEx \cite{ComplEx:conf/icml/TrouillonWRGB16} extends the representation of entities and relations to the complex vector space, providing better modeling of asymmetric relations. SimplE \cite{SimplE:conf/nips/Kazemi018}, based on the canonical polyadic decomposition, represents each entity and relation with two vectors, thereby increasing the correlation between head and tail entities. Additionally, some methods model relations in hypercomplex spaces to learn representations with more geometric features \cite{QuatE:conf/nips/0007TYL19},\cite{QuatRE:conf/www/NguyenVNP22}.

Neural network-based methods apply deep learning techniques to model entity and relation representations, aiming to obtain embeddings with strong generalization ability and robustness. Among them, ConvE \cite{ConvE:conf/aaai/DettmersMS018} is based on convolutional neural networks, R-GCN \cite{R-GCN:conf/esws/SchlichtkrullKB18} and M2GNN \cite{M2GNN:conf/www/WangWSWNAXYC21} are based on graph neural networks, and A2N \cite{A2N:conf/acl/BansalJRM19} leverages attention mechanisms. Additionally, there are methods like MuRP \cite{MuRP:conf/nips/BalazevicAH19} and ROTH \cite{ROTH:conf/acl/ChamiWJSRR20} that model entity and relation representations in non-Euclidean spaces.
Although KGE methods have been widely applied in recent years, they face limitations in capturing the dynamic evolution of facts in real-world scenarios. To address this problem, research focuses on addressing challenges such as handling dynamic knowledge graphs and incorporating temporal information.

\subsection{Temporal Knowledge Graph Embedding}
Temporal Knowledge Graph Embedding (TKGE) methods aim to incorporate temporal information into the representation learning process of knowledge graphs. This incorporation allows modeling of temporal dynamics and the evolution of facts. These methods extend traditional KGE approaches by introducing temporal factors, such as timestamps or time intervals, associated with triples in the knowledge graph. TTransE \cite{TTransE:conf/www/LeblayC18} extends TransE by considering time embedding as an equivalent vector to entities and relations. TA-DistMult \cite{TA-DistMult:conf/emnlp/Garcia-DuranDN18} extends DistMult by training a recursive neural network with sequences of tokens representing time predicates and digits in timestamps. DE-SimplE \cite{DE-SimplE:conf/aaai/GoelKBP20} extends SimplE by incorporating a diachronic entity embedding function to provide representations for entities at any given timestamp. In addition, TComplex \cite{TComplex:conf/iclr/LacroixOU20} extends ComplEx, BoxTE \cite{BoxTE:conf/aaai/MessnerAC22} extends BoxE \cite{BoxE:conf/nips/AbboudCLS20}, ChronoR \cite{ChronoR:conf/aaai/SadeghianACW21} and RotateQVS \cite{RotateQVS:conf/acl/ChenWLL22} extend RotatE.

On the other hand, some methods focus on incorporating time information by leveraging deep learning to craft customized network architectures or optimization functions. HyTE \cite{HyTE:conf/emnlp/DasguptaRT18} represents timestamps as mutually independent hyperplanes, where entities and relations satisfy the TransE assumption. ATiSE \cite{ATiSE} leverages the additive time series decomposition to treat entities and relations as temporal data, decomposing them into trend, seasonal, and random components. TeRo \cite{TeRo:conf/coling/XuNAYL20} combines the RotatE and TransE models by representing the temporal evolution of entity embeddings as a rotation in the complex vector space, starting from the initial time to the current time. DyERNIE \cite{DyERNIE:conf/emnlp/HanCMT20} models entities on a mixed curvature manifold and defines the tangent vector of a given entity as the velocity of the entity's evolution over time, enabling the description of dynamic evolutionary processes of entities. By incorporating temporal information, TKGE methods enhance knowledge graph embedding to capture temporal dynamics, and support downstream applications in evolving real-world scenarios. Although both mainstream categories of TKGE methods have their respective advantages, there is currently limited work that combines the strengths of these two approaches, namely having interpretable time representations while also supporting rich inference patterns. Therefore, this paper proposes a novel TKGE method via entity boxes and polynomial decomposition of time that aims to bridge this gap.

\section{Methodology}
In this section, we introduce the proposed PTBox method. The architecture is shown in Figure \ref{fig:overall}. We first describe the employed notations and definitions in Section \ref{sec:pf}. Then, we present the model framework and the two modules of our method in Section \ref{sec:mstr} and \ref{sec:beer}, respectively. In addition, we discuss the parameter learning strategy in Section \ref{sec:meq}, and model properties in Section \ref{sec:amp}.

\subsection{Problem Formulation}\label{sec:pf}
Temporal knowledge graphs represent events or facts using quadruples $(h, r, t, \tau)$, where $h\in\mathcal{E}$ and $t\in\mathcal{E}$ represent the head and tail entities, $r\in\mathcal{R}$ represents the relations, $\tau\in\mathcal{T}$ represents the timestamps. Then, a TKG can be formulated as $\mathcal{G}\subseteq\mathcal{E}\times\mathcal{R}\times\mathcal{E}\times\mathcal{T}$. For example, \textit{the quadruple (JosephineTewson, wasBornIn, Hampstead, 1931-02-26) describes the fact that Josephine Tewson was born in Hampstead on February 26, 1931.} 
TKGE methods aim to complete knowledge graphs by leveraging link prediction task, which utilize a scoring function to predict missing head or tail entities in $\mathcal{G}$ within a specified temporal context. Typically, the score function is learned with the formula as $f:\mathcal{E}\times\mathcal{R}\times\mathcal{E}\times\mathcal{T}\rightarrow\mathbb{R}$, that assigns a score $s=f(h,r,t,\tau)$ to each quadruple, indicating the prediction that a particular quadruple corresponds to a true fact. Followed closely, a non-linearity, such as the logistic or sigmoid function, is often used to convert the score to a predicted probability $p=\sigma(s)\in[0,1]$ of the quadruple being true.

\subsection{Polynomial Decomposition based Temporal Representation}
\label{sec:mstr}
The incorporation of temporal information represents the primary difference between TKG and traditional KG methods. As a result, effectively modeling temporal embeddings becomes a critical task for TKGE. Although existing TKGE methods have successfully integrated temporal information with entity-relation triplets, a limited number of approaches directly mathematically model continuous time information. Therefore, we model the timestamp via polynomial decomposition-based representation (PTR) to learn interpretable representations of continuous time.

According to Weierstrass approximation theorem \cite{PINKUS20001}, a continuous function defined on a closed interval can be uniformly approximated by a polynomial function. Namely, 
\begin{equation}
f\in C[a,b],\forall\epsilon>0,\exists P_n\Rightarrow\forall x\in[a,b],|f-P_n|<\epsilon 
\end{equation}
where $P_n$ is the polynomial function used to uniformly approximate the continuous function $f$. Based on Stone-Weierstrass theorem \cite{stoneweierstrass}, topological space $\mathbb{R}$ is a Hausdroff space, and the Weierstrass approximation theorem is satisfied in this space. Accordingly, we assume that time information can be expressed as a nonlinear function on the closed interval $[0,1]$, and design a multi-layer perceptron with a sigmoid output layer to learn this function. Then, we can leverage the Bernstein polynomial to represent $P_n$, which can be formulated as:
\begin{equation}
P_n(f_\tau,x)=\sum_{k=0}^n{f(\frac{k}{n})\dbinom{n}{k}x^k(1-x)^{n-k}}
\end{equation}
where $f_\tau(x)$ denotes the function describing time information. $f_\tau$ scales the input timestamps to $[0,1]$. Given $n$, we convert $P_n(f_\tau,x)$ into the matrix form, which can be formulated as:
\begin{equation}
\mathcal{P}_\tau=P_n(f_\tau,x)=\boldsymbol{\alpha_\tau}\cdot\boldsymbol{X},
\label{eq:ftxm}
\end{equation}
where $\boldsymbol{\alpha_\tau}$ denotes the coefficient matrix and $\boldsymbol{X}$ denotes the polynomial matrix. Note that $\boldsymbol{X}\in\mathbb{R}^{k\times d}$ is the temporal basis tensor, which learns the basic meta-features of time information. Then, based on Equation\ref{eq:ftxm}, we can easily model the temporal representation of any given timestamp. Moreover, according to the representations we can further dynamically model entities and relations under temporal constraints.

\begin{table*}[t]
\begin{center}
\scalebox{0.9}{
\begin{tabular}{lc}
\toprule
Inference Pattern & Setting\\
\midrule
Symmetry: $r_1(e_1,e_2|\tau)\Rightarrow r_1(e_2,e_1|\tau)$ & $P_{r_1}(e_1|e_2)=P_{r_1}(e_2|e_1)\neq0$\\
Antisymmetry: $r_1(e_1,e_2|\tau)\Rightarrow\urcorner r_1(e_2,e_1|\tau)$ & $P_{r_1}(e_1|e_2)\neq0,P_{r_1}(e_2|e_1)=0$ \\
Inversion: $r_1(e_1,e_2|\tau)\Leftrightarrow r_2(e_2,e_1|\tau)$ & $P_{r_1}(e_1|e_2)=P_{r_2}(e_2|e_1)\neq0$ \\
Composition: $r_1(e_1,e_2|\tau)\land r_2(e_2,e_3|\tau)\Rightarrow r_3(e_1,e_3|\tau)$ & $P_{r_3}(e_1,e_2,e_3)\neq0$ \\
Hierarchy: $r_1(e_1,e_2|\tau)\Rightarrow r_2(e_1,e_2|\tau)$ & $P_{r_1,r_2}(e_1|e_2)\geq P_{r_1}(e_1|e_2)P_{r_2}(e_1|e_2)\neq 0$ \\
Intersection: $r_1(e_1,e_2|\tau)\land r_2(e_1,e_2|\tau)\Rightarrow r_3(e_1,e_2|\tau)$ & $P_{r_3}(e_1|e_2)\geq P_{r_1,r_2}(e_1|e_2)\neq0$ \\
Mutual exclusion: $r_1(e_1,e_2|\tau)\land r_2(e_1,e_2|\tau)\Rightarrow\bot$ & $P(\mathcal{B}(e_{1}^{r_1})\cap\mathcal{B}(e_{2}^{r_1}),\mathcal{B}(e_{1}^{r_2})\cap\mathcal{B}(e_{2}^{r_2}))=0$ \\
\bottomrule
\end{tabular}}
\end{center}
\caption{Inference patterns/generalized inference patterns captured by our PTBox with fixed timestamp $\tau$.}\label{tab:pattern}
\end{table*}
\subsection{Box Embedding based Entity Representation}
\label{sec:beer}
Temporal knowledge graphs inherently contain rich geometric structural information. Geometric embedding methods possess the natural ability to represent transitive asymmetric relations via containment. Among these methods, box embeddings represent objects as $n$-dimensional hyperrectangles and exhibit closure under intersection. This characteristic makes box embeddings well-suited for capturing complex relationships within knowledge graphs. Hence, we model the entities via box embedding based entity representations (BER) to enhance the representation and reasoning capabilities of our PTBox model.

In our PTBox, every entity $e_i\in\mathcal{E}$ is represented by a $n$-dimensional axis-aligned hyperrectangle (namely box) $Box(e_i)\subseteq\mathbb{R}^d$, which can be viewed as a lattice, a special poset. Each box is represented by a pair of vectors, which correspond to the maximum coordinates $e_i^M$ and minimum coordinates $e_i^m$ of the box, respectively. Considering the entities in the knowledge graph as a non-strict partial order set, we can then define the relationship between $e_i$ and $e_j$ by inclusion of boxes as follows:
\begin{equation}
\resizebox{0.95\hsize}{!}{$
\begin{aligned}
e_i\vee e_j&=\prod_k{[\min(e_k^{m,i},e_k^{m,j}), \max(e_k^{M,i},e_k^{M,j})]},\\
e_i\wedge e_j&=\begin{cases}
    \bot,&\mbox{if } e_i, e_j \mbox{ disjoint}\\
    \prod_k{[\max(e_k^{m,i},e_k^{m,j}), \min(e_k^{M,i},e_k^{M,j})]},&\mbox{otherwise}
\end{cases}
\end{aligned}$}
\label{eq:pol}
\end{equation}
where $\vee$, $\wedge$, and $\bot$ denote partial order relations, $\vee$ is the smallest enclosing box, $\wedge$ is the intersecting box, and $\bot$ is the empty box. $\min(\cdot)$ and $\max(\cdot)$ are functions used to calculate the minimum and maximum values, respectively. Further, following the viewpoint proposed by Vilnis et al. \cite{Query2Box:conf/acl/McCallumVLM18}, we can interpret the volume of a box as a non-normalized probability. Therefore, utilizing inclusion-exclusion with set intersection over $Box(\cdot)$, the joint probability and conditional probability of relationships in the knowledge graph can be easily calculated. 
\begin{equation}
\begin{aligned}
P(e_i,e_j,e_k)=&Vol(Box(e_i)\cap Box(e_j)\cap Box(e_k)),\\
P(e_i|e_j)=&\frac{Vol(Box(e_i)\cap Box(e_j))}{Vol(Box(e_j))},
\end{aligned}
\label{eq:prob}
\end{equation}
where $P$ denotes the probability function, $Vol(\cdot)$ denotes the volume function of the box, and $\cap$ denotes the intersection operator between boxes.

However, the parameter settings of box embeddings result in equivalent probability distributions, rendering conventional gradient-based deep learning optimization algorithms impractical for learning \cite{Query2Box:conf/acl/McCallumVLM18}. To mitigate this, we employ Gumbel boxes proposed by Dasgupta et al. \cite{GumbelBox:conf/nips/DasguptaBZV0M20,chen2021probabilistic} to model our box embeddings. The maximum and minimum coordinates of Gumbel boxes follow the Gumbel distribution, then the boxes can be formulated as:
\begin{equation}
\begin{aligned}
Box(e)&=\prod_{i=1}^{d}{[e_i^m,e_i^M]},\\
e_i^m&\sim\mbox{MaxGumbel}(\mu_i^m,\beta),\\
e_i^M&\sim\mbox{MinGumbel}(\mu_i^M,\beta),
\end{aligned}
\label{eq:box}
\end{equation} 
where $\mu$ is a location parameter and $\beta$ is a scale parameter. The mean and variance of Gumbel distribution are $\mu+\gamma\beta$ and $\frac{\pi^2}{6}\beta^2$, where $\gamma$ is the Euler–Mascheroni constant. Gumbel distribution as generalized extreme value distribution is min and max stable, keeping the Gumbel boxes closed under intersection. Hence, the approximation of volume in Gumbel boxes can be formulated as:
\begin{equation}
\resizebox{0.88\hsize}{!}{$
\mathbb{E}[Vol(Box(e))]\approx\prod_{i=1}^{d}\beta\log{(1+\exp(\frac{\mu_i^M-\mu_i^m}{\beta}-2\gamma))}$}.
\label{eq:vol}
\end{equation}

\subsection{Modeling and Evaluation of Quadruples}\label{sec:meq}
As mentioned previously, given a quadruple $(h,r,t,\tau)$, our method models head and tail entities as two Gumbel boxes $Box(e_h)$, $Box(e_t)$. Timestamp $\tau$ is modeled as a temporal projection $\mathcal{P}_\tau$. Then, the evolutionary dynamics of entities and relations over time can be formulated as:
\begin{equation}
\resizebox{0.9\hsize}{!}{$
\begin{aligned}
e_h^{\prime}=&\mathcal{P}_\tau(e_h;W)=Box(e_h)+(W^TBox(e_h))W,\\
e_t^{\prime}=&\mathcal{P}_\tau(e_t;W)=Box(e_t)+(W^TBox(e_t))W,\\
r_t^{\prime}=&\mathcal{P}_\tau(r_t;W)=r_t+(W^Tr_t)W,
\end{aligned}$}
\label{eq:tmap}
\end{equation}
where $W$ denotes the weight of temporal projection, $e_h^\prime$ and $e_t^\prime$ denote the evolutionary representations of head and tail entities, $r_t^\prime$ denotes the evolutionary representation of relation. Note that the mapped entity representations remain Gumbel boxes.

Furthermore, we consider each relation $r$ as an affine transformation $T_r\subseteq\mathbb{R}^{2\times d}$ acting on entity boxes, where $T_r[0]$ represents the translation and $T_r[1]$ represents the scaling. Given a entity $e$, the relation transformation can be formulated as:
\begin{equation}
\begin{aligned}
e^t=&f_r^t(e|T_r)=Box(e)+T_r[0],\\
e^s=&f_r^s(e|T_r)=Box(e)\odot T_r[1],
\end{aligned}
\label{eq:rtrans}
\end{equation}
where $e^t$ denotes the translation of $e$, $e^s$ denotes the scaling of $e$, and $\odot$ is the Hadamard product. To simplify the notations, we use $f_r^t(\cdot)$ and $f_r^s(\cdot)$ to denote the two transformation operations of relation $r$. The composition of these functions satisfies the properties of an Abelian group, allowing them to act on Gumbel boxes while preserving the relationships and structure of the boxes. Consequently, given the quadruple $(h,r,t,\tau)$, we define a scoring function as the volume intersection between two new boxes formed by the evolved boxes of $h$ and $t$ under the transformation of $r$ at $\tau$. This can be formulated as:
\begin{equation}
\resizebox{1.0\hsize}{!}{$
\mathcal{S}(h,r,t,\tau)=\frac{\mathbb{E}[Vol(f_r^s\circ f_r^t(\mathcal{P}_\tau(e_h;W))\cap f_r^s\circ f_r^t(\mathcal{P}_\tau(e_t;W)))]}{\mathbb{E}[Vol(f_r^s\circ f_r^t(\mathcal{P}_\tau(e_t;W)))]}$.}
\end{equation}

\subsection{Analysis of Model Properties}\label{sec:amp}
We analyze the representation power and inductive capacity of PTBox. The conclusion of our analysis indicates that PTBox is locally identifiable and can capture rich inference patterns and higher-arity relations. We additionally analyze the complexity of PTBox, and prove that it runs in time $O(d)$ and space $O((|E|+|R|+K)d)$, where $|E|$ and $|R|$ are the maximal quantity.

\begin{table}
\begin{center}
\scalebox{1.0}{
\begin{tabular}{*{3}c}
\toprule
Dataset & YAGO11k & WikiData\\
\midrule
Entities & 10,623 & 500 \\
Relations & 10 & 24 \\
Time Span & -453$\sim$2844 & 1479$\sim$2018 \\
Train & 16,408 & 32,497 \\
Validation & 2,050 & 4,062 \\
Test & 2,051 & 4,062 \\
\bottomrule
\end{tabular}}
\caption{Statistics of two experimented datasets.}\label{tab:dataset}
\end{center}
\end{table}

\noindent\textbf{Local Identifiability.} The local identifiability of a model is used to measure whether its parameters are sensitive to local features. Assuming a set of parameters $\Omega$ is local identifiable if, for all $\theta\in\Omega$, there exists $N(\theta)$, a neighborhood of $\theta$, such that for all $\theta^\prime\in N(\theta),L(x|\theta^\prime)\neq L(x|\theta)$. According to Vilnis et al.'s observations \cite{Query2Box:conf/acl/McCallumVLM18}, the parameter space of probability box embeddings has large degrees of freedom, which leads to the lack of local identifiability. Although this property implies that the model's parameters are not overly influenced by local variations or noise in the data, the lack of local identifiability poses significant challenges in the training and optimization of the model. Our approach tackles this problem by employing the Gumbel distribution to represent box embeddings.
To effectively mitigate the aforementioned problems with optimization, we retain the uncertainty over the box intersections and make all parameters contribute to the data likelihood in an appropriate manner. Moreover, computing the volume using conditional probabilities in Equation\ref{eq:prob} can further alleviate the unboundedness problem in the base measure space \cite{GumbelBox:conf/nips/DasguptaBZV0M20}.

\noindent\textbf{Inference Patterns.} We study the inductive capacity of PTBox in terms of common inference patterns appearing in the TKGE literature, and details as shown in Table\ref{tab:pattern}. Inference patterns are important for downstream tasks of knowledge graphs, and jointly capturing multiple inference patterns is meaningful but challenging. Our PTBox captures all generalized inference patterns given in Table\ref{tab:pattern} through box configurations. For example, PTBox inherently supports probability and intersection rules. Symmetry can be captured by ensuring the conditional probabilities $P_{r_1}(e_1|e_2)$ and $P_{r_1}(e_2|e_1)$ exist and are equal. Composition can be captured by ensuring the joint probability $P_{r_3}(e_1,e_2,e_3)$ exists. Mutual exclusion is captured by disjointness between the intersection boxes under relations $r_1$ and $r_2$, respectively. Compared to earlier methods, TransE \cite{TransE:conf/nips/BordesUGWY13} fails to capture symmetry and composition, RotatE \cite{RotatE:conf/iclr/SunDNT19} fails to capture hierarchy, and ComplEx \cite{ComplEx:conf/icml/TrouillonWRGB16} fails to capture composition and intersection. Clearly, our method captures more diverse inference patterns and exhibits stronger inductive capability. In addition, based on the configuration of PTBox, we compute the evolved representations of relations at different timestamps, allowing us to capture cross-time inference patterns, as mentioned in \cite{BoxTE:conf/aaai/MessnerAC22}. 

\noindent\textbf{Runtime and Space Complexity.} In terms of runtime complexity, for any quadruple $(h,r,t,\tau)$, we firstly compute $d$-dimensional temporal embedding through polynomial decomposition. Then, we multiply relations to obtain two $d$-dimensional relation embeddings, and finally compute the evolved entity embeddings using multiplication and addition operations. The volume function runs in $O(d)$ for every box. In terms of space complexity, PTBox stores two $d$-dimensional vectors for each entity box, two $d$-dimensional vectors for each relation, and a $K\times d$ matrix for time. Hence, for a KG with $|E|$ entities and $|R|$ relations, PTBox requires $(|E| + |R| + K)d$ parameters.

\begin{table*}[t]
\begin{center}
\scalebox{1.0}{
\begin{tabular}{l*{6}{c}}
\toprule
\multicolumn{1}{c}{\multirow{2}*{Model}} & \multicolumn{3}{c}{YAGO11k} & \multicolumn{3}{c}{WikiData} \\
& MRR & Hits@3 & Hits@10 & MRR & Hits@3 & Hits@10 \\
\midrule
TransE & 0.100 & 0.138 & 0.244 & 0.178 & 0.192 & 0.339 \\
DistMult & 0.158 & 0.161 & 0.268 & 0.222 & 0.238 & 0.460 \\
RotatE & 0.167 & 0.167 & 0.305 & 0.116 & 0.236 & 0.461 \\
QuatE & 0.164 & 0.148 & 0.270 & 0.125 & 0.243 & 0.416 \\
\hline
TTransE & 0.108 & 0.150 & 0.251 & 0.172 & 0.184 & 0.329 \\
HyTE & 0.105 & 0.143 & 0.272 & 0.180 & 0.197 & 0.333 \\
TA-DistMult & 0.161 & 0.171 & 0.292 & 0.218 & 0.232 & 0.447 \\
ATiSE & 0.170 & 0.171 & 0.288 & 0.280 & 0.317 & 0.481 \\
TeRo & \underline{0.187} & 0.197 & 0.319 & {\bf0.299} & \underline{0.329} & \underline{0.507} \\
RotateQVS & {\bf0.189} & \underline{0.199} & \underline{0.323} & - & - & - \\
\hline
PTBox & 0.162 & {\bf0.222} & {\bf0.347} & \underline{0.290} & {\bf0.331} & {\bf0.527} \\
\bottomrule
\end{tabular}}
\caption{Link prediction results on YAGO11k, WikiData for our proposed and baseline methods. The best results are marked in bold.}\label{tab:sota}
\end{center}
\end{table*}

\section{Experiments}
We evaluate the performance of our proposed PTBox model on two popular TKG benchmarks.
\subsection{Experimental Setup}
\textbf{Datasets.} We use two well-known datasets for evaluation, namely, YAGO11k, and WikiData. YAGO11k and WikiData are temporal subgraphs extracted from YAGO3 and Wikipedia, respectively. As a subset of YAGO3, YAGO11k incorporates information from multiple sources, including Wikipedia, WordNet, and GeoNames, ensuring its richness and reliability. In this paper, the WikiData is Wikipedia12k proposed by HyTE \cite{HyTE:conf/emnlp/DasguptaRT18}. Similar to YAGO11k, Wikipedia12k contains the facts involving time intervals. The detailed statistics of the datasets are presented in Table \ref{tab:dataset}.

\noindent\textbf{Evaluation Metrics.} We utilize the link prediction task to evaluate the effectiveness of the PTBox model. We employ classic evaluation metrics, which include mean rank (MR), mean reciprocal rank (MRR) and hits at 1/3/10 (Hits@1/3/10). These metrics all represent the rankings of missing ground-truth entities in the prediction results. For each query, we report the mean results of both the subject and object entity prediction tasks. 

\noindent\textbf{Implementation Details.} We implemented our PTBox model in PyTorch and trained the model on a GPU (RTX 3090). We configured the parameters based on the MRR and Hits@10 performance achieved by the model on the validation set. For polynomial decomposition based temporal representation, the order of temporal polynomial $k$ is set to 20. For box embedding based entity representation, the embedding dimension $d$ is set to 128, the distribution of box embeddings follows $Gumbel(0.01,1)$ and $Gumbel(-0.1, -0.001)$. We train our model by Adam optimizer, and set the learning rate as 0.0001 for all datasets.

\subsection{Results of TKGE on Link Prediction}
In this section, we compare the performance of our proposed PTBox with that of static and dynamic methods based on the TKG link prediction task.

\noindent\textbf{Baseline Models.} We compare our approach with several state-of-the-art (SOTA) approaches, including static KGE methods and dynamic TKGE methods. Among them, TransE \cite{TransE:conf/nips/BordesUGWY13}, DistMult \cite{DistMult:journals/corr/YangYHGD14a}, RotatE \cite{RotatE:conf/iclr/SunDNT19}, and QuatE \cite{QuatE:conf/nips/0007TYL19} are static KGE methods, which focus on modeling triples of static facts. TTransE \cite{TTransE:conf/www/LeblayC18}, HyTE \cite{HyTE:conf/emnlp/DasguptaRT18}, TA-DistMult \cite{TA-DistMult:conf/emnlp/Garcia-DuranDN18}, ATiSE \cite{ATiSE}, TeRo \cite{TeRo:conf/coling/XuNAYL20}, and RotateQVS \cite{RotateQVS:conf/acl/ChenWLL22} are dynamic TKGE methods, which focus on modeling quadruples of temporal facts.

\noindent\textbf{Experimental Results.} We present the performance of different models on temporal datasets in Table \ref{tab:sota}. From the results compared to the baseline models, our method consistently achieves performance improvements across all datasets. We observe that PTBox outperforms RotateQVS in terms of Hits@3 and Hits@10 on YAGO11k. PTBox also outperforms TeRo in terms of Hits@3 and Hits@10, and is competitive with TeRo in terms of MRR on WikiData. Specifically, our method achieves an MRR of 16.2\%, Hits@3 of 22.2\%, and Hits@10 of 34.7\% on the YAGO11k, and an MRR of 29.0\%, Hits@3 of 33.1\%, and Hits@10 of 52.7\% on the WikiData, respectively.

Table \ref{tab:sota} lists the link prediction results on On YAGO11k and WikiData where time annotations in facts are time intervals. Compared with static KGE methods, our PTBox outperforms RotatE by 5.5\% regarding Hits@3, and by 4.2\% regarding Hits@10 on YAGO11k, respectively. Our PTBox also outperforms DistMult by 6.8\% regarding MRR, outperforms QuatE by 8.8\% regarding Hits@3, and outperforms RotatE by 6.6\% regarding Hits@10 on WikiData, respectively. It means that temporal information can effectively enhance the performance of knowledge completion and reasoning by supplementing the facts. Compared with temporal KGE methods, our PTBox outperforms RotateQVS by 2.3\% regarding Hits@3, and by 2.4\% regarding Hits@10 on YAGO11k, respectively. It means that our proposed strategy of modeling time through polynomial decomposition can effectively capture temporal information, while providing good embedding representations for temporal knowledge graphs. On the other hand, our PTBox underperforms on the MRR but outperforms on the Hits@3 and Hits@10 compared to both RotateQVS and TeRo. This indicates that our model has limitations in top-1 accuracy, but has higher recall compared to theirs. In addition, both of these methods fix the representation of timestamps in the dataset during the training phase, and they cannot model unseen timestamps as flexibly as our PTBox does.

\begin{table}
\begin{center}
\scalebox{1.0}{
\begin{tabular}{l*{4}{c}}
\toprule
\multicolumn{1}{c}{\multirow{2}*{Model}} & \multicolumn{2}{c}{YAGO11k} & \multicolumn{2}{c}{WikiData} \\
& MR & Hits@1 & MR & Hits@1 \\
\midrule
TransE & 1.70 & 0.784 & 1.35 & 0.884 \\
TransH & 1.53 & 0.761 & 1.40 & 0.881 \\
HolE & 2.57 & 0.693 & 2.23 & 0.840 \\
\hline
t-TransE & 1.66 & 0.755 & 1.97 & 0.742 \\
HyTE & 1.23 & 0.812 & 1.13 & 0.926 \\
\hline
PTBox & {\bf1.12} & {\bf0.896} & {\bf1.12} & {\bf0.934} \\
\bottomrule
\end{tabular}}
\caption{Relation prediction results on YAGO11k and WikiData for our proposed and baseline methods. The best results are marked in bold.}\label{tab:sota_rp}
\end{center}
\end{table}

\begin{table*}[t]
\begin{center}
\scalebox{1.0}{
\begin{tabular}{*{12}{c}}
\toprule
\multicolumn{1}{c}{\multirow{2}*{PTR}} & \multicolumn{1}{c}{\multirow{2}*{BER}} & \multicolumn{3}{c}{YAGO11k} & \multicolumn{3}{c}{WikiData} \\
&& MRR & Hits@3 & Hits@10 & MRR & Hits@3 & Hits@10 \\
\midrule
&& 0.105 & 0.143 & 0.272 & 0.180 & 0.197 & 0.333 \\
$\checkmark$ && 0.137 & 0.174 & 0.313 & 0.259 & 0.278 & 0.478 \\
& $\checkmark$ & 0.127 & 0.180 & 0.280 & 0.253 & 0.281 & 0.426 \\
$\checkmark$ & $\checkmark$ & {\bf0.162} & {\bf0.222} & {\bf0.347} & {\bf0.290} & {\bf0.331} & {\bf0.527} \\
\bottomrule
\end{tabular}}
\caption{Ablation study on YAGO11k and WikiData for our proposed two strategies under link prediction task. The best results are marked in bold.}\label{tab:abl1}
\end{center}
\end{table*}

\begin{table}[t]
\begin{center}
\scalebox{1.0}{
\begin{tabular}{*{7}{c}}
\toprule
\multicolumn{1}{c}{\multirow{2}*{Mode}} & \multicolumn{2}{c}{YAGO11k} & \multicolumn{2}{c}{WikiData} \\
& MRR & Hits@10 & MRR & Hits@10 \\
\midrule
$\mathcal{P}_\tau(e)$ & 0.127 & 0.289 & 0.267 & 0.485 \\
$\mathcal{P}_\tau(r)$ & {\bf0.162} & {\bf0.347} & {\bf0.290} & {\bf0.527} \\
$\mathcal{P}_\tau(e,r)$ & 0.135 & 0.311 & 0.277 & 0.503 \\
\bottomrule
\end{tabular}}
\caption{Ablation study on YAGO11k and WikiData for examining different evolutionary patterns of knowledge graph triples over time. The best results are marked in bold.}\label{tab:abl2}
\end{center}
\end{table}

\subsection{Results of TKGE on Relation Prediction}
To further analyze the performance of our model, we conduct relation prediction experiments to examine whether temporal information is beneficial in resolving ambiguities among relations. Detailed results are shown in Table \ref{tab:sota_rp}.

Based on the experimental results, the three static methods (TransE, TransH, and HolE \cite{HolE:conf/aaai/NickelRP16}) demonstrate inferior performance compared to the two dynamic methods (HyTE and our PTBox) on both the YAGO11k and WikiData. It means that time information surely contributes to relation disambiguation. Although the t-TransE \cite{tTransE16:conf/emnlp/JiangLGSLCS16} is a temporal model, it does not directly model temporal information. Instead, it achieves implicit temporal fusion through relation ordering. As a result, compared to the other two methods, it does not fully leverage temporal information to resolve ambiguities among relations. Different from HyTE can only model times present in the training set, our approach utilizes the Weierstrass approximation theorem (as outlined in Section \ref{sec:mstr}) to learn a coefficient matrix for any given time, and then we can easily model the temporal representation of any timestamp by combining with the temporal basis tensor (as outlined in Equation.\ref{eq:pol}). It means that our proposed polynomial based temporal representation offers more flexibility in modeling time, which enhances the performance of relation prediction. Consequently, our method has achieved SOTA performance on both datasets. Specifically, compared with HyTE, we can increase Hits@1 from 81.2\% to 89.6\% on YAGO11k, and from 92.6\% to 93.4\% on WikiData, respectively.

\subsection{Ablation Study}
In this section, we conduct ablation study to evaluate the effectiveness and necessity of our proposed components on two datasets. As mentioned in Section \ref{sec:mstr} and \ref{sec:beer}, we propose two strategies to model time and entities, namely PTR and BER. Therefore, we compare the impact of different strategies on performance across all datasets, and the results are reported in Table \ref{tab:abl1}. Moreover, we compare different evolutionary patterns of the knowledge graph over time, as shown in Table \ref{tab:abl2}.

\noindent\textbf{PTR and BER.} As shown in Table \ref{tab:abl1}, we employ HyTE as the baseline and report the experimental results of PTR and BER when activated independently and jointly. The experimental results demonstrate that both the proposed PTR and BER contribute to enhancing the performance of KGE on link prediction tasks when compared to the baseline model. In particular, their joint application leads to significant improvements across all datasets. According to the results in the second row, our PTR strategy can boost the performance of link prediction by modeling temporal information more effectively than baseline model. According to the results in the third row, our BER strategy also benefits in improving model performance. This shows that leveraging geometric modeling and intersection operation can cover rich reasoning patterns. Comparing the second and third rows of Table \ref{tab:abl1}, we can observe that the PTR strategy improves performance slightly better than BER. This means that effective modeling of time information is more important in the TKGE methods.

\begin{table*}
\begin{center}
\scalebox{0.9}{
\begin{tabular}{*{3}c}
\toprule
Test quadruples & HyTE & Ours\\
\midrule
Katie\_Holmes, ?, Tom\_Cruise, [2006, 2012] & \textbf{isMarriedTo}, hasWonPrize & \textbf{isMarriedTo}, created \\
Tricia\_Devereaux, ?, Illinois, [1975, 1975] & diedIn, \textbf{wasBornIn} & \textbf{wasBornIn}, diedIn \\
Jeremy\_Lloyd, ?, London, [2014, 2014] & wasBornIn, \textbf{diedIn} & \textbf{diedIn}, wasBornIn \\
Will\_Haining, ?, Fleetwood\_Town\_F.C., [2011, -] & isMarriedTo, \textbf{playsFor} & \textbf{playsFor}, isMarriedTo \\
Bob\_Hope, ?, Toluca\_Lake,\_Los\_Angeles, [2003, 2003] & isMarriedTo, hasWonPrize & isMarriedTo, \textbf{diedIn} \\
\bottomrule
\end{tabular}}
\caption{Case study of qualitative analysis on relation prediction. The order of prediction is in descending order. Correct one is in \textbf{bold}.}\label{tab:qa_case}
\end{center}
\end{table*}

\begin{figure}
\centering
\includegraphics[scale=0.6]{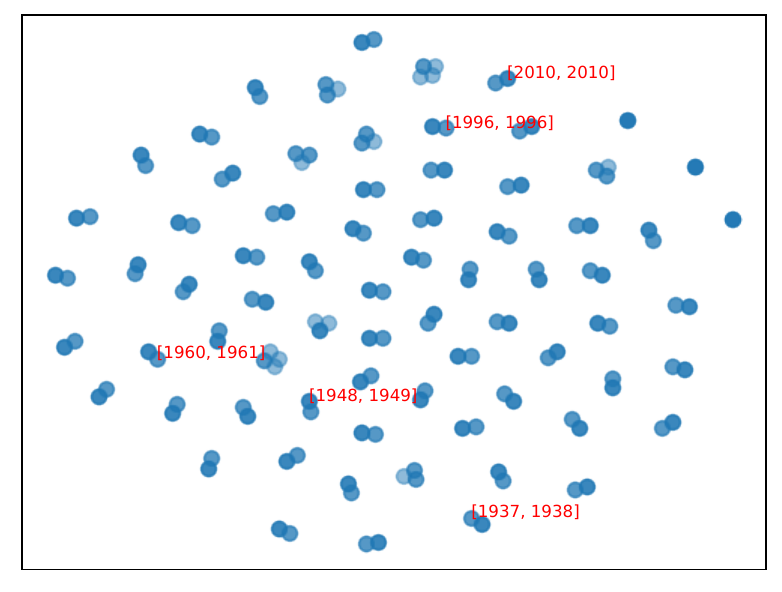}
\caption{Visualization of polynomial decomposition based temporal representations on YAGO11k.}
\label{fig:time_emb}
\end{figure}

\noindent\textbf{Different Evolutionary Patterns.} To explore the impact of different evolutionary patterns on the model, we summarize the experimental results of three different patterns in Table \ref{tab:abl2}. Among them, $\mathcal{P}_\tau(e)$ represents time acting on entities, $\mathcal{P}_\tau(r)$ represents time acting on relations, and $\mathcal{P}_\tau(e,r)$ represents time simultaneously acting on both entities and relations. Compared to entities, the types of relations are fewer and their representations in the feature space are relatively sparse. Therefore, the evolution of relations over time can produce more significant differences. We believe that this view is also consistent with real-world scenarios, where individuals maintain independence and the changes over time are the relationships between individuals.

\subsection{Qualitative Analysis} 
We conducted two qualitative analysis experiments to intuitively demonstrate the performance of our model, including the case study of relation prediction with results shown in Table \ref{tab:qa_case} and the visualization of temporal representations with results illustrated in Figure \ref{fig:time_emb}.

Table \ref{tab:qa_case} presents a comparative analysis of predictions made on some samples from the YAGO11k test set by our method and HyTE. Overall, our approach demonstrates superior accuracy in relation prediction compared to HyTE. This is particularly evident in the case of the easily confusable relations $diedIn$ and $wasBornIn$, where our method consistently predicts the ground truth accurately. Furthermore, for the fact \textit{(Bob Hope, diedIn, Toluca Lake Los Angeles, [2003, 2003])}, unlike HyTE, which fails to deliver the correct result, our method achieves a hit within the top 2 predictions.

Evaluation metric for continuous modelling of time is challenging to quantify, but we can employ the same strategy as HyTE, visualizing the distribution of temporal embeddings in the vector space. From Figure \ref{fig:time_emb}, we can observe that timestamps with close time intervals are close to each other in vector space, and vice versa. This indicates that our proposed polynomial decomposition based temporal representation is capable of effectively modeling temporal information in a continuous manner.

\section{Conclusion}
In this paper, we propose an innovative TKGE method based on polynomial approximation for modeling arbitrary time information, namely PTBox. Our main contributions lies in polynomial decomposition-based temporal representation and box embedding-based entity representation. To enhance the performance of the TKGE, we focus on improving the capabilities to continuously model arbitrary time information and infer under temporal constraints. Our method decomposes timestamp by polynomial approximation theory to flexibly represent time information. Furthermore, to capture complex geometric structures and learn rich inference patterns, we model entities by box embeddings and define each relation as a transformation on the head and tail entity boxes. Theoretically, our proposed PTBox can encode arbitrary time information or even unseen timestamps, while capturing higher-arity relations of the knowledge base. Extensive experiments on real-world datasets demonstrate the effectiveness of our method.

\section{Acknowledgements}
This research was supported by the National Science and Technology Major Project (2022ZD0119204), National Key Research and Development Program of China (2020AAA0109701), National Science Fund for Distinguished Young Scholars (62125601).

\section{References}
\bibliographystyle{lrec-coling2024-natbib}
\bibliography{../references}

\begin{thebibliography}{38}
\expandafter\ifx\csname natexlab\endcsname\relax\def\natexlab#1{#1}\fi

\bibitem[{Abboud et~al.(2020)Abboud, Ceylan, Lukasiewicz, and
  Salvatori}]{BoxE:conf/nips/AbboudCLS20}
Ralph Abboud, {\.I}smail~{\.I}lkan Ceylan, Thomas Lukasiewicz, and Tommaso
  Salvatori. 2020.
\newblock Boxe: {A} box embedding model for knowledge base completion.
\newblock In \emph{Advances in Neural Information Processing Systems}, pages
  9649--9661.

\bibitem[{Balazevic et~al.(2019)Balazevic, Allen, and
  Hospedales}]{MuRP:conf/nips/BalazevicAH19}
Ivana Balazevic, Carl Allen, and Timothy~M. Hospedales. 2019.
\newblock Multi-relational poincar{\'{e}} graph embeddings.
\newblock In \emph{Advances in Neural Information Processing Systems}, pages
  4465--4475.

\bibitem[{Bansal et~al.(2019)Bansal, Juan, Ravi, and
  McCallum}]{A2N:conf/acl/BansalJRM19}
Trapit Bansal, Da{-}Cheng Juan, Sujith Ravi, and Andrew McCallum. 2019.
\newblock {A2N:} attending to neighbors for knowledge graph inference.
\newblock In \emph{Proceedings of the Conference of the Association for
  Computational Linguistics}, pages 4387--4392.

\bibitem[{Bordes et~al.(2013)Bordes, Usunier, Garc{\'{\i}}a{-}Dur{\'{a}}n,
  Weston, and Yakhnenko}]{TransE:conf/nips/BordesUGWY13}
Antoine Bordes, Nicolas Usunier, Alberto Garc{\'{\i}}a{-}Dur{\'{a}}n, Jason
  Weston, and Oksana Yakhnenko. 2013.
\newblock Translating embeddings for modeling multi-relational data.
\newblock In \emph{Advances in Neural Information Processing Systems}, pages
  2787--2795.

\bibitem[{Chami et~al.(2020)Chami, Wolf, Juan, Sala, Ravi, and
  R{\'{e}}}]{ROTH:conf/acl/ChamiWJSRR20}
Ines Chami, Adva Wolf, Da{-}Cheng Juan, Frederic Sala, Sujith Ravi, and
  Christopher R{\'{e}}. 2020.
\newblock Low-dimensional hyperbolic knowledge graph embeddings.
\newblock In \emph{Proceedings of the Annual Meeting of the Association for
  Computational Linguistics}, pages 6901--6914.

\bibitem[{Chen et~al.(2022)Chen, Wang, Li, and
  Li}]{RotateQVS:conf/acl/ChenWLL22}
Kai Chen, Ye~Wang, Yitong Li, and Aiping Li. 2022.
\newblock Rotateqvs: Representing temporal information as rotations in
  quaternion vector space for temporal knowledge graph completion.
\newblock In \emph{Proceedings of the Annual Meeting of the Association for
  Computational Linguistics}, pages 5843--5857.

\bibitem[{Chen et~al.(2021)Chen, Boratko, Chen, Dasgupta, Li, and
  Mccallum}]{chen2021probabilistic}
Xuelu Chen, Michael Boratko, Muhao Chen, Shib~Sankar Dasgupta, Xiang~Lorraine
  Li, and Andrew Mccallum. 2021.
\newblock Probabilistic box embeddings for uncertain knowledge graph reasoning.
\newblock In \emph{Proceedings of the Conference of the North American Chapter
  of the Association for Computational Linguistics: Human Language
  Technologies}, pages 882--893.

\bibitem[{Cotter(1990)}]{stoneweierstrass}
N.E. Cotter. 1990.
\newblock The stone-weierstrass theorem and its application to neural networks.
\newblock \emph{IEEE Transactions on Neural Networks}, 1(4):290--295.

\bibitem[{Dasgupta et~al.(2020)Dasgupta, Boratko, Zhang, Vilnis, Li, and
  McCallum}]{GumbelBox:conf/nips/DasguptaBZV0M20}
Shib~Sankar Dasgupta, Michael Boratko, Dongxu Zhang, Luke Vilnis, Xiang Li, and
  Andrew McCallum. 2020.
\newblock Improving local identifiability in probabilistic box embeddings.
\newblock In \emph{Advances in Neural Information Processing Systems}, pages
  182--192.

\bibitem[{Dasgupta et~al.(2018)Dasgupta, Ray, and
  Talukdar}]{HyTE:conf/emnlp/DasguptaRT18}
Shib~Sankar Dasgupta, Swayambhu~Nath Ray, and Partha~P. Talukdar. 2018.
\newblock Hyte: Hyperplane-based temporally aware knowledge graph embedding.
\newblock In \emph{Proceedings of the Conference on Empirical Methods in
  Natural Language Processing}, pages 2001--2011.

\bibitem[{Dettmers et~al.(2018)Dettmers, Minervini, Stenetorp, and
  Riedel}]{ConvE:conf/aaai/DettmersMS018}
Tim Dettmers, Pasquale Minervini, Pontus Stenetorp, and Sebastian Riedel. 2018.
\newblock Convolutional 2d knowledge graph embeddings.
\newblock In \emph{Proceedings of the {AAAI} Conference on Artificial
  Intelligence}, pages 1811--1818.

\bibitem[{Ebisu and Ichise(2018)}]{TorusE:conf/aaai/EbisuI18}
Takuma Ebisu and Ryutaro Ichise. 2018.
\newblock Toruse: Knowledge graph embedding on a lie group.
\newblock In \emph{Proceedings of the {AAAI} Conference on Artificial
  Intelligence}, pages 1819--1826.

\bibitem[{Garc{\'{\i}}a{-}Dur{\'{a}}n et~al.(2018)Garc{\'{\i}}a{-}Dur{\'{a}}n,
  Dumancic, and Niepert}]{TA-DistMult:conf/emnlp/Garcia-DuranDN18}
Alberto Garc{\'{\i}}a{-}Dur{\'{a}}n, Sebastijan Dumancic, and Mathias Niepert.
  2018.
\newblock Learning sequence encoders for temporal knowledge graph completion.
\newblock In \emph{Proceedings of the Conference on Empirical Methods in
  Natural Language Processing}, pages 4816--4821.

\bibitem[{Goel et~al.(2020)Goel, Kazemi, Brubaker, and
  Poupart}]{DE-SimplE:conf/aaai/GoelKBP20}
Rishab Goel, Seyed~Mehran Kazemi, Marcus~A. Brubaker, and Pascal Poupart. 2020.
\newblock Diachronic embedding for temporal knowledge graph completion.
\newblock In \emph{Proceedings of the {AAAI} Conference on Artificial
  Intelligence}, pages 3988--3995.

\bibitem[{Han et~al.(2020)Han, Chen, Ma, and
  Tresp}]{DyERNIE:conf/emnlp/HanCMT20}
Zhen Han, Peng Chen, Yunpu Ma, and Volker Tresp. 2020.
\newblock Dyernie: Dynamic evolution of riemannian manifold embeddings for
  temporal knowledge graph completion.
\newblock In \emph{Proceedings of the Conference on Empirical Methods in
  Natural Language Processing}, pages 7301--7316.

\bibitem[{Hu et~al.(2022)Hu, Zhao, Huang, and Huang}]{hu2022global}
Shiyu Hu, Xin Zhao, Lianghua Huang, and Kaiqi Huang. 2022.
\newblock Global instance tracking: Locating target more like humans.
\newblock \emph{IEEE Transactions on Pattern Analysis and Machine
  Intelligence}, 45(1):576--592.

\bibitem[{Jiang et~al.(2016)Jiang, Liu, Ge, Sha, Li, Chang, and
  Sui}]{tTransE16:conf/emnlp/JiangLGSLCS16}
Tingsong Jiang, Tianyu Liu, Tao Ge, Lei Sha, Sujian Li, Baobao Chang, and
  Zhifang Sui. 2016.
\newblock Encoding temporal information for time-aware link prediction.
\newblock In \emph{Proceedings of the Conference on Empirical Methods in
  Natural Language Processing}, pages 2350--2354.

\bibitem[{Kazemi and Poole(2018)}]{SimplE:conf/nips/Kazemi018}
Seyed~Mehran Kazemi and David Poole. 2018.
\newblock Simple embedding for link prediction in knowledge graphs.
\newblock In \emph{Advances in Neural Information Processing Systems}, pages
  4289--4300.

\bibitem[{Lacroix et~al.(2020)Lacroix, Obozinski, and
  Usunier}]{TComplex:conf/iclr/LacroixOU20}
Timoth{\'{e}}e Lacroix, Guillaume Obozinski, and Nicolas Usunier. 2020.
\newblock Tensor decompositions for temporal knowledge base completion.
\newblock In \emph{Proceedings of the International Conference on Learning
  Representations}.

\bibitem[{Leblay and Chekol(2018)}]{TTransE:conf/www/LeblayC18}
Julien Leblay and Melisachew~Wudage Chekol. 2018.
\newblock Deriving validity time in knowledge graph.
\newblock In \emph{Companion of the International World Wide Web Conferences},
  pages 1771--1776.

\bibitem[{Lv et~al.(2018)Lv, Hou, Li, and Liu}]{TransC:conf/emnlp/LvHLL18}
Xin Lv, Lei Hou, Juanzi Li, and Zhiyuan Liu. 2018.
\newblock Differentiating concepts and instances for knowledge graph embedding.
\newblock In \emph{Proceedings of the Conference on Empirical Methods in
  Natural Language Processing}, pages 1971--1979.

\bibitem[{Messner et~al.(2022)Messner, Abboud, and
  Ceylan}]{BoxTE:conf/aaai/MessnerAC22}
Johannes Messner, Ralph Abboud, and {\.I}smail~{\.I}lkan Ceylan. 2022.
\newblock Temporal knowledge graph completion using box embeddings.
\newblock In \emph{Proceedings of the {AAAI} Conference on Artificial
  Intelligence}, pages 7779--7787.

\bibitem[{Nguyen et~al.(2022)Nguyen, Vu, Nguyen, and
  Phung}]{QuatRE:conf/www/NguyenVNP22}
Dai~Quoc Nguyen, Thanh Vu, Tu~Dinh Nguyen, and Dinh~Q. Phung. 2022.
\newblock Quatre: Relation-aware quaternions for knowledge graph embeddings.
\newblock In \emph{Companion of the International World Wide Web Conferences},
  pages 189--192.

\bibitem[{Nickel et~al.(2016)Nickel, Rosasco, and
  Poggio}]{HolE:conf/aaai/NickelRP16}
Maximilian Nickel, Lorenzo Rosasco, and Tomaso~A. Poggio. 2016.
\newblock Holographic embeddings of knowledge graphs.
\newblock In \emph{Proceedings of the {AAAI} Conference on Artificial
  Intelligence}, pages 1955--1961.

\bibitem[{Nickel et~al.(2011)Nickel, Tresp, and
  Kriegel}]{RESCAL:conf/icml/NickelTK11}
Maximilian Nickel, Volker Tresp, and Hans{-}Peter Kriegel. 2011.
\newblock A three-way model for collective learning on multi-relational data.
\newblock In \emph{Proceedings of the International Conference on Machine
  Learning}, pages 809--816.

\bibitem[{Pinkus(2000)}]{PINKUS20001}
Allan Pinkus. 2000.
\newblock Weierstrass and approximation theory.
\newblock \emph{Journal of Approximation Theory}, 107(1):1--66.

\bibitem[{Sadeghian et~al.(2021)Sadeghian, Armandpour, Colas, and
  Wang}]{ChronoR:conf/aaai/SadeghianACW21}
Ali Sadeghian, Mohammadreza Armandpour, Anthony Colas, and Daisy~Zhe Wang.
  2021.
\newblock Chronor: Rotation based temporal knowledge graph embedding.
\newblock In \emph{Proceedings of the {AAAI} Conference on Artificial
  Intelligence}, pages 6471--6479.

\bibitem[{Schlichtkrull et~al.(2018)Schlichtkrull, Kipf, Bloem, van~den Berg,
  Titov, and Welling}]{R-GCN:conf/esws/SchlichtkrullKB18}
Michael~Sejr Schlichtkrull, Thomas~N. Kipf, Peter Bloem, Rianne van~den Berg,
  Ivan Titov, and Max Welling. 2018.
\newblock Modeling relational data with graph convolutional networks.
\newblock In \emph{Proceedings of the European Semantic Web Conference}, volume
  10843, pages 593--607.

\bibitem[{Sun et~al.(2019)Sun, Deng, Nie, and Tang}]{RotatE:conf/iclr/SunDNT19}
Zhiqing Sun, Zhi{-}Hong Deng, Jian{-}Yun Nie, and Jian Tang. 2019.
\newblock Rotate: Knowledge graph embedding by relational rotation in complex
  space.
\newblock In \emph{Proceedings of the International Conference on Learning
  Representations}.

\bibitem[{Trouillon et~al.(2016)Trouillon, Welbl, Riedel, Gaussier, and
  Bouchard}]{ComplEx:conf/icml/TrouillonWRGB16}
Th{\'{e}}o Trouillon, Johannes Welbl, Sebastian Riedel, {\'{E}}ric Gaussier,
  and Guillaume Bouchard. 2016.
\newblock Complex embeddings for simple link prediction.
\newblock In \emph{Proceedings of the International Conference on Machine
  Learning}, volume~48, pages 2071--2080.

\bibitem[{Vilnis et~al.(2018)Vilnis, Li, Murty, and
  McCallum}]{Query2Box:conf/acl/McCallumVLM18}
Luke Vilnis, Xiang Li, Shikhar Murty, and Andrew McCallum. 2018.
\newblock Probabilistic embedding of knowledge graphs with box lattice
  measures.
\newblock In \emph{Proceedings of the Annual Meeting of the Association for
  Computational Linguistics}, pages 263--272.

\bibitem[{Wang and Cheng(2018)}]{TransN:conf/sigir/WangC18}
Chun{-}Chih Wang and Pu{-}Jen Cheng. 2018.
\newblock Translating representations of knowledge graphs with neighbors.
\newblock In \emph{Proceedings of the International {ACM} {SIGIR} Conference on
  Research {\&} Development in Information Retrieval}, pages 917--920.

\bibitem[{Wang et~al.(2021)Wang, Wei, dos Santos, Wang, Nallapati, Arnold,
  Xiang, Yu, and Cruz}]{M2GNN:conf/www/WangWSWNAXYC21}
Shen Wang, Xiaokai Wei, C{\'{\i}}cero~Nogueira dos Santos, Zhiguo Wang, Ramesh
  Nallapati, Andrew~O. Arnold, Bing Xiang, Philip~S. Yu, and Isabel~F. Cruz.
  2021.
\newblock Mixed-curvature multi-relational graph neural network for knowledge
  graph completion.
\newblock In \emph{Proceedings of the International World Wide Web Conference},
  pages 1761--1771.

\bibitem[{Wang et~al.(2014)Wang, Zhang, Feng, and
  Chen}]{TransH:conf/aaai/WangZFC14}
Zhen Wang, Jianwen Zhang, Jianlin Feng, and Zheng Chen. 2014.
\newblock Knowledge graph embedding by translating on hyperplanes.
\newblock In \emph{Proceedings of the {AAAI} Conference on Artificial
  Intelligence}, pages 1112--1119.

\bibitem[{Xu et~al.(2020{\natexlab{a}})Xu, Nayyeri, Alkhoury, Yazdi, and
  Lehmann}]{TeRo:conf/coling/XuNAYL20}
Chengjin Xu, Mojtaba Nayyeri, Fouad Alkhoury, Hamed~Shariat Yazdi, and Jens
  Lehmann. 2020{\natexlab{a}}.
\newblock Tero: {A} time-aware knowledge graph embedding via temporal rotation.
\newblock In \emph{Proceedings of the International Conference on Computational
  Linguistics}, pages 1583--1593.

\bibitem[{Xu et~al.(2020{\natexlab{b}})Xu, Nayyeri, Alkhoury, Yazdi, and
  Lehmann}]{ATiSE}
Chenjin Xu, Mojtaba Nayyeri, Fouad Alkhoury, Hamed Yazdi, and Jens Lehmann.
  2020{\natexlab{b}}.
\newblock Temporal knowledge graph completion based on time series gaussian
  embedding.
\newblock In \emph{Proceedings of the International Semantic Web Conference},
  pages 654--671.

\bibitem[{Yang et~al.(2015)Yang, Yih, He, Gao, and
  Deng}]{DistMult:journals/corr/YangYHGD14a}
Bishan Yang, Wen{-}tau Yih, Xiaodong He, Jianfeng Gao, and Li~Deng. 2015.
\newblock Embedding entities and relations for learning and inference in
  knowledge bases.
\newblock In \emph{Proceedings of the International Conference on Learning
  Representations}.

\bibitem[{Zhang et~al.(2019)Zhang, Tay, Yao, and
  Liu}]{QuatE:conf/nips/0007TYL19}
Shuai Zhang, Yi~Tay, Lina Yao, and Qi~Liu. 2019.
\newblock Quaternion knowledge graph embeddings.
\newblock In \emph{Advances in Neural Information Processing Systems}, pages
  2731--2741.

\end{thebibliography}

\bibliographystylelanguageresource{lrec-coling2024-natbib}
\bibliographylanguageresource{languageresource}

\end{document}